\def\year{2020}\relax
\documentclass[letterpaper]{article} 
\usepackage{aaai20}  
\usepackage{times}  
\usepackage{helvet} 
\usepackage{courier}  
\usepackage[hyphens]{url}  
\usepackage{graphicx} 
\usepackage{graphicx}  
\usepackage{amsmath}
\usepackage{comment}
\usepackage{algorithm}
\usepackage{algorithmic}
\usepackage[switch]{lineno}  
\title{Extending Label Smoothing Regularization with Self-Knowledge Distillation}

\author{Ji-Yue Wang\textsuperscript{\rm 1,3}\textbf{, Pei Zhang}\textsuperscript{\rm 2,4}\textbf{, Wen-feng Pang}\textsuperscript{\rm 1,5}\textbf{, Jie Li}\textsuperscript{\rm 1,6}\\ 
	\textsuperscript{\rm 1}School of Electronic and Information Engineering, South China University of Technology, GuangZhou, China\\
	\textsuperscript{\rm 2}School of Computer Science, Northwestern Polytechnical University, Xi'an, China\\ 
\textsuperscript{\rm 3}jiyuewang@outlook.com, \textsuperscript{\rm 4}cszhangpei@mail.nwpu.edu.cn, \textsuperscript{\rm 5}wenfengpang@gmail.com \textsuperscript{\rm 6}eejli@scut.edu.cn\\ 
}

\begin{document}
	\maketitle
	
	\begin{abstract}
		Inspired by the strong correlation between the Label Smoothing Regularization(LSR) and Knowledge distillation(KD), we propose an algorithm LsrKD for training boost by extending the LSR method to the KD regime and applying a softer temperature. Then we improve the LsrKD by a Teacher Correction(TC) method, which manually sets a constant larger proportion for the right class in the uniform distribution teacher. To further improve the performance of LsrKD, we develop a self-distillation method named Memory-replay Knowledge Distillation (MrKD) that provides a knowledgeable teacher to replace the uniform distribution one in LsrKD. The MrKD method penalizes the KD loss between the current model's output distributions and its copies' on the training trajectory. By preventing the model learning so far from its historical output distribution space, MrKD can stabilize the learning and find a more robust minimum. Our experiments show that LsrKD can improve LSR performance consistently at no cost, especially on several deep neural networks where LSR is ineffectual. Also, MrKD can significantly improve single model training. The experiment results confirm that the TC can help LsrKD and MrKD to boost training, especially on the networks they are failed. Overall, LsrKD, MrKD, and their TC variants are comparable to or outperform the LSR method, suggesting the broad applicability of these KD methods. 
		
	\end{abstract}

	\section{Introduction}
	
	\noindent Deep learning has been a story of booms of success, yet, as the network becomes deeper and wider \cite {He2016,Iandola2014,chen2017dual}, the model consumes more and more computational resources. There is a trend to use light models with fewer parameters to save memory and accelerate learning and inferring speed \cite {Howard2017,Sandler2018,Howard2019,Ma2018} . With carefully designed supernet space and model searching strategy, Neural Architecture Search(NAS) techniques \cite {Liu2018,Tan2019} can find proper models to fit different requirements (flops, memory). Besides that, efforts are delivered to extract a small model from powerful large ones, e.g., pruning \cite {Li2016}, binarisation \cite {Rastegari2016}, encoding \cite {Han2015}, and knowledge distillation \cite {Hinton2015}.
	
	Knowledge Distillation (KD) \cite {Hinton2015} compressed the knowledge from the teacher model, which is a larger model or a set of multiple models, to a single small student model. The knowledge is transferred from a pre-trained teacher model to a student model with a Kullback-Leibler(KL) divergence loss between their output probabilities. In addition to its many application in model compression, KD is also used to boost network training with multiple models that have identical architecture \cite {Furlanello2019,Zhang2018} or single model self-distillation \cite {Zhang2019,Yun2020,Hendrycks2019}. 
	
	\begin{figure}[t]
		\centering
		\includegraphics[width=0.9\columnwidth]{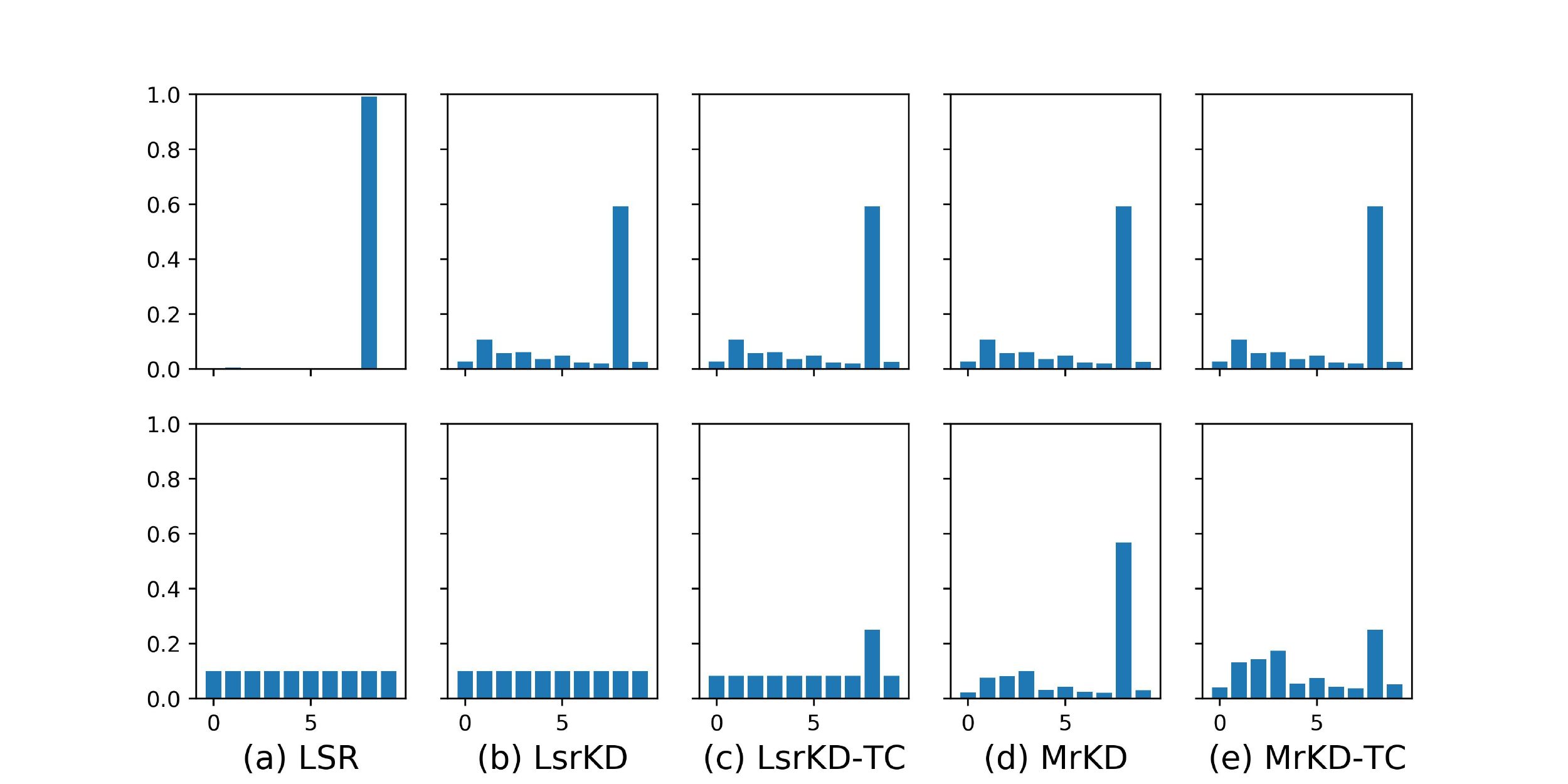} 
		\caption{Distribution examples of our proposed methods. Upper: current student model distribution with different $\tau$ used for Knowledge Distillation, Lower: Corresponding teacher distributions obtained by different methods }
		\label{fig}	
	\end{figure}
	
	In this paper, we consider helping training with KD methods. Based on the observation that knowledge distillation can be interpreted as a regularization method and inspired by the strong correlation between the KD method and LSR, we reformulate the LSR method in KD expression as \cite {Yuan2019}. Here, LSR can be regarded as a special case of KD with a uniform distribution teacher, and its softmax temperature equals to 1 (Fig. 1(a)). Considering the importance of the softer temperature in KD \cite {Hinton2015}, we generalize the LSR method to LsrKD with a hype-parameter temperature $\tau$ instead of 1 (Fig. 1(b)). Then, to make the uniform distribution teacher more informative, we propose a Teacher Correction (TC) method that manually sets a constant larger proportion $\gamma$ for the correct class in the uniform distribution (Fig. 1(c)).  
	
	However, the handcrafted LsrKD with TC will only go so far; it is still lifeless and can only provide limited guidance. Hence, we further propose a self-distillation method called Memory Replay Knowledge Distillation(MrKD) to obtain a more knowledgeable teacher. Note that no extra model \cite {Furlanello2019,Zhang2018} or structure \cite {Zhang2019} is required in our strategy: the knowledge is distilled from the model backup during the training trajectory (Fig. 1(d)). Our method is based on the assumption that a student can be improved by reflecting on his own experience. The network backup's parameter  $\hat{ \theta }$ is updated to current model parameters $\theta$ every $\kappa$ steps during the training procedure.
	This model update strategy is rarely used in conventional supervised learning but is a common practice in deep reinforcement learning methods \cite {Mnih2015} for the target network renewal. Besides the conventional supervised learning loss, the KL-divergence loss between the current and the backup model will also be penalized for regularizing the model to a more flat result.

	It seems counterintuitive that a student can learn from a teacher worse than himself. Nevertheless, an observation in KD techniques is that a much larger teacher model usually does not end up with a better student than a medium-scale teacher due to the capacity mismatching problem \cite {Cho2019}. Also, the Deep Mutual Learning method (DML) \cite {Zhang2018} shows that distillation with peers is better than training a student with a pre-trained teacher statically; even the worse peer can help its opponent learning. \cite {Yuan2019} reveals the correlation between LSR and KD and suggests that a manually designed teacher can also help training. The above observations suggest that in addition to inducing the similarity of category information, the observations above show that KD's efficiency is due to the regularization effect to a large extent. 
	
	\cite {Mandt2017} showed that stochastic gradient descent(SGD) with a constant learning rate simulates a sampling from a Gaussian distribution centered at the loss's minimum. Following this idea, we regard the whole SGD training procedure as a Markov Chain trajectory sampled from a dynamic transition distribution parameterized by learning rate, mini-batch size, sampling order, and the model weight initialization. 
	When the SGD learning is proceeding, although the weight update gradually and the model's output are changing slowly, the model's latent distribution can differ from each other in detail. The similar but diverse distribution of the model backup $\kappa$ steps ago can an be informational reference for the model to achieve a more general minimum. 
	
	Our experiments demonstrate that LSR has limited improvement when the network is deep or complicated, whereas LsrKd and MrKD can help training on different network architectures consistently. Although the backup outputs are worse than the current model, the 'dark knowledge' \cite {Hinton2015} that the backup offered in MrKD improves training further than LsrKD. Furthermore, we observe that Teacher Correction can help the LsrKD and MrKD when they lose efficacy on complicated networks like WRN-28-10 \cite {Zagoruyko2016}.
	
	The contributions of this work are summarized as follows:
	
	\begin{itemize}
		\item By applying a softer temperature to Label Smoothing Regularization in Knowledge Distillation form,  LsrKD provides a reliable substitution to LSR.
		\item We propose a self-distillation method Memory Replay Knowledge Distillation(MrKD) which utilizes the training trajectory model backup as a teacher. MrKD can offer dark knowledge conveniently to improve the generalization of training.
		\item We introduce Teacher Correction to improve LsrKD and MrKD with a trustworthy teacher. Our results demonstrate that the proposed methods of LsrKD, MrKD, and their TC variants outperform the original LSR. Our methods are easy to implement in neural networks with little modification and training procedures; thus, they can be widely used tools.
	\end{itemize} 
	
	\section{Related work}
	
	\noindent \textbf{Multiple model KD for boost training.} Born Again Network (BAN) \cite {Furlanello2019} trained students parameterized identically to their teacher, and the outperform their teachers significantly. The authors use the pre-trained model as a teacher to train a student and set the trained student as the teacher for the next training iteration. However, the recurrent distillation of BAN requires high computation and storage costs.
	
	The Deep Mutual Learning method \cite {Zhang2018} used an ensemble of students to learn collaboratively and showed that the mutual learning strategy performs better than the static teacher-student mode. Furthermore, a larger teacher net can also benefit from this mutual learning. However, aggregating students' logits to form an ensemble teacher restrain student peers' diversity, thus limit the effectiveness of online learning \cite {Wang2020}. Their work shows an essential characteristic of KD: the teacher is not necessarily perfect or accurate. That is, an intermediate output from a teacher that matching the student's training procedure, is comparable to a precise result from a pre-trained teacher.  \cite {Jin2019} also confirmed a similar idea. 

	It is difficult to learn from a larger or more precise teacher. This is the initial motivation for us to investigate the efficacy of the uniform distribution teacher in LsrKD and the  self-distillation MrKD method.
	
	\noindent \textbf{Single model KD.} \cite {Zhang2019} proposed a self-distillation method that divides a single network into several sections connected with additional bottleneck and fully connected layers to constitute multiple classifiers. Then the knowledge in the deepest classifier of the network is squeezed into the shallower ones. The study of self-distillation is promising; they claimed that the teacher branch improves the shallower sections' learning features. \cite {Luan2019} deepened the shallower section's bottleneck classifier and applied mutual learning distillation instead of the teacher-student method and achieve better performance. This improvement of MSD indicates that the self-distillation method can be regarded as a DML method of four peers with different low-level weight sharing. We evaluate four-model DML directly and found comparable results. Except with fewer parameters, this self-distillation method \cite {Zhang2019} can also be regarded as a multi-model KD method as DML.
	These network remodeling or model ensembling methods \cite {Song2018,Zhang2019,Zhu2018} have the limitation of generalization and flexibility.
	
	Furthermore, KD loss can also regularize the model output consistency of similar training samples, such as augmented data and original data \cite {Hendrycks2019}, or samples belong to the same classes \cite {Yun2020}. However, the former method relies on the efficacy of the augmentation method, and the latter needs a carefully designed training procedure.
	
	\noindent \textbf{The efficacy of KD.} \cite {Cho2019} shows that the reason for larger models not making good teachers is capacity mismatching: small students are unable to mimic large teachers. They applied early stopping to both teacher pre-training and student mimic learning. This method proved to be effective and improve the ImageNet dataset classification remarkably. However, the problem is just alleviated rather than solved. With this technique, the best teacher they have found for ResNet18 \cite {He2016} on the ImageNet dataset among all ResNet families (from ResNet18 to ResNet152) is ResNet32.
	
	Experimentally and theoretically, \cite {Yuan2019} found that KD can be interpreted as a regularization method, and they revealed the relation between KD and LSR. Their proposed Teacher-free KD (Tf-KD) method first designed a teacher as our TC method and then applied high temperature ($\tau$ $\geq$ 20)) on KD loss. The hyper-parameters in Tf-KD is model dependent and hard to tune. Our LsrKD method emphasizes that a proper soft temperature is more critical than the hand-crafted teacher. LsrKD is tuned on one model and performs consistently on all models. Without conducting the LSR in KD form, \cite {Xu2020} replace the uniform distribution in the LSR method directly with the output of a teacher model pre-trained on the ImageNet dataset and help the training on CIFAR100.
	
	\noindent \textbf{KD with historical models}
	To alleviate the capacity mismatching problem \cite {Cho2019}, \cite {Mirzadeh2019} introduces multi-step KD, which uses an intermediate-sized model (teacher assistant) to bridge the gap between the student and teacher. Route Constrained Optimization(RCO) \cite {Jin2019} supervises the student model with some anchor points selected from the route in parameter space that the teacher pass by, instead of the converged teacher model. Our MrKD method extracts the anchor points progressively from the student itself during the training.  
	
	Inspired by the fact that averaging model weights over training steps tends to find a flatter solution  \cite {Izmailov2018}, the Mean Teacher \cite {Tarvainen2017} method distilled the knowledge from a teacher that averages successive steps model weights and improved the performance of semi-supervised tasks. 
	\cite {Xu2020} fine-tuned the BERT model in Natural Language Processing problems by distilling the knowledge of the averaged weight parameter of $\kappa$ recent steps. 
	The recent time steps historical model KD can help semi-supervised learning or model fine-tuning but scarcely improve common classification problems. Our work on MrKD reveals that the model backups far away from the current training can regularize supervised learning effectively.
	
	\section{Methodology}
	In this section, we present our proposed methods. First, we formulate the Knowledge Distillation method (Subsection 3.1). Next, by reformulating the LSR technique in KD form, we propose the LsrKD method (Subsection 3.2) and amend it with the Teacher Correction method (Subsection 3.3). Finally, to employ a more informative teacher than uniform distribution in LsrKD, we explain how MrKD uses the model's historical backups to guide the current model in Subsection 3.4.
	\subsection{Knowledge Distillation}
	We consider a standard image classification problem. Given a training dataset D = \{(x$^{i}$, y$^{i}$)\}$_{i=1}^N$, where \textit{x$^{i}$} is the \textit{i$_{th}$} sample from M classes and \textit{y$^{i}$}=\{1, 2,..., M\} is the corresponding label of sample \textit{x$^{i}$}, the parameters $\theta$ of a deep neural network(DNN) that best fit to the dataset need to be determined. 
	
	The softmax function is employed to calculate the \textit{m$_{th}$} class probability from a given model:
	\begin{equation}
	q_m(\tau)=\frac{exp(z_m/\tau)}{\sum_{i=1}^{M}{exp(z_i/\tau)}}
	\end{equation}
	Here \textit {z$_{m}$} is the \textit{m$_{th}$} logit output of the model’s fully connected layer. $\tau$ indicates the temperature of softmax distribution normally set to 1 in traditional cross-entropy loss but greater than 1 in knowledge distillation loss \cite{Hinton2015}. A larger $\tau$ means a softer probability distribution that reveals more detail than a hard softmax output ($\tau$=1). 
	
	Firstly, we introduce the standard cross-entropy loss of one sample for M-class classification:
	\begin{equation}
	L_{CE}(p,q(1))=-\sum_{m=1}^{M}{p_{m}log(q_m(1))}
	\end{equation}
	Where \textit{p$_{m}$} is the \textit{m$_{th}$} element of one-hot label vector \textit{p}. Note that the temperature $\tau$ is set to 1.
	
	In the KD method, a pre-trained teacher will output a corresponding logit $\hat {z}$. 
	To transfer the knowledge form a teacher model to the student, Kullback Leibler (KL) Divergence between their output probabilities is used:
	\begin{equation}
	L_{KL}(\hat{q}(\tau)||q(\tau)) =\sum_{m=1}^{M}{\hat{q}_{m}(\tau)log(\frac{\hat{q}_{m}(\tau)}{q_{m}(\tau)})}
	\end{equation}
	Here the temperature $\tau$ is a hyper-parameter need to be tuned, and the $\hat{q}_{m}$ is obtained by Eq. (1) with $\hat {z_m}$.
	During training, the KD method calculates the sum of two losses above with a hyper-parameter $\alpha$:
	\begin{equation}
	L_{KD}=(1-\alpha)*L_{CE}(p,q(1))+ \alpha*{\tau}^{2}L_{KL}(\hat{q}(\tau)||q(\tau))
	\end{equation}
	Where  ${\tau}^{2} $ is a factor in ensuring that the relative contribution of the ground-truth label and teacher output distribution remains roughly unchanged \cite{Hinton2015}.
	
	\subsection{Label Smoothing Regularization with Knowledge Distillation}
	In LSR, for a training example with one-hot label vector \textit{p}, \cite{Szegedy2016} replaced the \textit{p} as \textit{p$^{'}$}:
	\begin{equation}
	p^{'} =(1-\alpha)*p+ \alpha*u
	\end{equation}
	where \textit{u} is a uniform distribution. As \cite {Yuan2019} showed, the cross-entropy loss of LSR can be written as a KD loss, similar to Eq. (4):
	\begin{equation}
	\begin{aligned}
	L_{LSR}(p,q){} & = L_{CE}(p^{'},q(1))\\
	& = (1-\alpha) L_{CE}(p,q(1))+ \alpha L_{KL}(u,q(1))
	\end{aligned}
	\end{equation}
	which means that LSR can be regarded as a special case of KD with a uniform distribution teacher and $\tau$ = 1.
	\cite{Hinton2015} has shown that a soft temperature is critical for KD methods perform well, so we extend the KL loss to a generalized form and put forward the LsrKD method:
	\begin{equation}
	\begin{aligned}
	L_{LsrKD}(p,q) = {} &(1-\alpha)*L_{CE}(p,q(1))\\
	& +\alpha*L_{KL}(u,q(\tau))
	\end{aligned}
	\end{equation}
	In the KD method, a factor $\tau^{2}$ on the KL loss is applied to stabilize the back-prop gradient while $\tau$ changing. Here, we analyze the KL loss gradient with $\tau$ in Eq. (7) briefly as in \cite{Hinton2015}:
	\begin{equation}
	\begin{aligned}
	\frac{\partial L_{KL}(u,q(\tau))}{\partial z_m} {} &= \frac{1}{\tau} (q_m(\tau)-u_m)\\
	&= \frac{1}{\tau}(\frac{exp(z_m/\tau)}{\sum_{i=1}^{M}{exp(z_i/\tau)}}-u_m)
	\end{aligned}
	\end{equation}
	If the temperature is high to the logits' magnitude, and the logits of model output have been zero-meaned. Eq. (9) simplifies to:
	\begin{equation}
	\begin{aligned}
	\frac{\partial L_{KL}(u,q(\tau))}{\partial z_m} {} &\approx \frac{1}{\tau}(\frac{1+z_m/\tau}{M}-u_m)\\
	& =  \frac{1}{M\tau}(z_m/\tau-(Mu_m-1)) \\
	& =  \frac{z_m}{M\tau^2}\\
	\end{aligned}
	\end{equation}
	From the conduction above, in the high-temperature limit, the KL loss regularization will penalize large and confidence logit values \cite{Pereyra2017}. Although the gradient is proportional to 1/$\tau^2$ above, we observe that a factor $\tau$ in amending the gradient stability instead of $\tau^{2}$ for the KL loss achieves better results if the temperature is not that large. Then our final LsrKD loss can be written as:
	\begin{equation}
	\begin{aligned}
	L_{LsrKD}(p,q) = {} &(1-\alpha)*L_{CE}(p,q(1))\\
	& +\alpha*{\tau} L_{KL}(u,q(\tau))
	\end{aligned}
	\end{equation}
	\subsection{Teacher Correction}
	
	In the LsrKD method, the uniform distribution teacher \textit{u} can be substituted as below \cite {Yuan2019}:
	\begin{equation}
	q^{'}_{m} = 
	\left\{   
	\begin{array}{ll}
	\gamma, & m=c\\
	\frac{1-\gamma}{M-1}, & m\neq c
	\end{array} 
	\right.
	\end{equation}
	c is the correct label, and $\gamma$ is the probability of class c. We call this Teacher Correction(TC).
	Then the LsrKD-TC loss is:
	\begin{equation}
	L_{LsrKD-TC}(p,q) = (1-\alpha)L_{CE}(p,q(1))+\alpha{\tau} L_{KL}(q^{'},q(\tau))
	\end{equation}

	\subsection{Memory Replay Knowledge Distillation}
	\subsubsection{Formulation}
	
	In every $\kappa$ steps during the training, the backup model weights $\hat{\theta}$ will be updated to the current model $\theta$. The identical structure model with parameter $\hat{\theta}$ is used as a teacher in Eq. (3). Thus, our MrKD loss is:
	\begin{equation}
	L_{MrKD}=(1-\alpha)*L_{CE}(p,q(1))+ \alpha*{\tau}^{2}L_{KL}(\hat{q}(\tau)||q(\tau))
	\end{equation}
	The proposed MrKD method can extend to \textit{n} memory copies $\hat{\theta}_{1}$,...,$\hat{\theta}_{n}$, with $\kappa$ training steps interval. The KL loss in Eq. (13) is extend to:
	\begin{equation}
	L_{KL}(\hat{q}_1,...,\hat{q}_n||q) = \frac{1}{n} \sum_{i=1}^{n}{L_{KL}(\hat{q}_i||q) }
	\end{equation}
	\begin{algorithm}
		\caption{Memory Replay Knowledge Distillation}
		\begin{algorithmic} 
			\STATE \textbf{Require:} training set D, learning rate $\lambda_t$, kd loss ratio $\alpha$, copy step interval $\kappa$, copy amount \textit{n}, temperature $\tau$, total training steps T
			\STATE \textbf{Initilize:} model parameters    $\theta$, $\hat{\theta}_{1}$, ..., $\hat{\theta}_{n}$
			\FOR {\textit{t}=1,...,T}
			\IF {$ \;(t \bmod\;\kappa)$ == 0}
			
			\FOR {\textit{i}=\textit{n}, ...,2}
			\STATE $\hat{\theta}_{i}$ := $\hat{\theta}_{i-1}$
			\ENDFOR
			\STATE $\hat{\theta}_{1}$ := $\theta$
			\ENDIF
			\STATE Sample a mini-batch of data \textit{d} form D
			\STATE Feed \textit{d} to networks and get logits \textit{z}, $\hat{z}_{1}$, ..., $\hat{z}_{n}$
			\STATE Compute the predictions \textit{q}(1), \textit{q}($\tau$) , $\hat{q}_{1}$($\tau$), ..., $\hat{q}_{n}$($\tau$) by Eq. (1)
			\STATE Compute loss $L_{total}$ ($\theta$) by Eq. (13)
			\STATE Update $\theta$ with stochastic gradient descent:
			\begin{equation}
			\theta := \theta - \lambda_t\frac{\partial L_{MrKD}}{\partial \theta}
			\end{equation}
			\ENDFOR
		\end{algorithmic}
	\end{algorithm}

	\subsubsection{Algorithm}
	The training procedure is shown in Algorithm 1. With every $\kappa$ steps, all the model copies' parameters $\hat{\theta}_{1}$,...,$\hat{\theta}_{n}$, are updated recursively.
	In each step, a mini-batch d is sampled and fed into the current model and its copies. With the models' logit outputs \textit{z}, $\hat{z}_{1}$, ..., $\hat{z}_{n}$, we can get the probabilities of mini-batch d by Eq. (1). Then the loss is calculated by Eq. (13). Finally, the current model parameters $\theta$ is updated by SGD as Eq. (15). Note that this algorithm can benefit from multiple GPUs Training. If n+1 GPUs are available, where n is the number of copies, distributed forward pass can be implemented for n+1 models, then the training time will be identical to the standard training method. 
	\subsubsection{TC extension}
	We can also extend the MrKD method by TC, the memory replay output $\hat q_{m}(\tau)$ can be replaced with:
	\begin{equation}
	\hat q^{'}_{m}(\tau) = 
	\left\{   
	\begin{array}{ll}
	\gamma , & m=c\\
	(1-\gamma)*\frac{\hat q_{m}(\tau)}{1-\hat q_{c}(\tau) }, & m\neq c
	\end{array} 
	\right.
	\end{equation}
	the probability of correct class c is set to $\gamma$, and the rest of the classes will share the rest 1-$\gamma$ with proportion to their original probability.
	Finally, the loss of MrKD-TC method is:
	\begin{equation}
	L_{MrKD-TC}=(1-\alpha)L_{CE}(p,q(1))+ \alpha{\tau} L_{KL}(\hat{q}^{'}(\tau)||q(\tau))
	\end{equation}

	\section{Experiments}

	\subsection{Dataset}
	Three datasets are used in our experiments. The \textbf{CIFAR10} and \textbf{CIFAR100} datasets consist of $32\times32$ color images in 10 and 100 classes respectively. Both are split into 50 000 training images and 10 000 testing images.	The \textbf{CINIC10} dataset is an extended version of CIFAR10. It contains all images from CIFAR10 and derives 210 000 images downsampled to 32x32 from the ImageNet dataset. For all the three datasets above, a random horizontal flip and crop with 4 pixels zero-padding are applied for training. 

	\subsection{Implementation Details}
	We implement all networks and training procedures in PyTorch and conduct all experiments on a single NVIDIA TITAN RTX GPU.
	The networks used in our experiments are all implemented strictly as their official papers for tiny image datasets (CIFAR-10, CIFAR100, and CINIC-10) without modification, including ResNet, PreActResNet, WideResNet, ResNeXt, etc. 
	
For all runs, including the baselines, we train a total epoch of 200, with a weight decay of 0.0005, a momentum of 0.9, a batch size of 128, and an initial learning rate of 0.1 that decreases to 0.0001 with cosine annealing. We record the last epoch results of 4 runs for all presented results because we found that choosing the best epoch results prone to benefit unstable and oscillating configurations. Then we calculate the mean and standard deviation of the 4 results. 
\subsection{Hyper-parameters Tuning}

There are three hyper-parameters for LsrKD and LsrKD-TC we need to tune: the KD loss weight $\alpha$, temperature $\tau$, and the TC factor $ \gamma$. For the MrKD method, there are two extra parameters: model backup update frequency $\kappa$ and copy amount \textit{n}.
We evaluate these parameters cursorily on the CIFAR100 dataset following the setting in subsection 4.2 with the ranges below:
\begin{itemize}
	\item $\alpha$: \{0.01, 0.05, 0.1, 0.25, 0.5, 0.75, 0.9\}
	\item $\tau$: \{ 1, 2, 3, 4, 5, 8, 12, 20\}
	\item $\gamma$: \{ 0.01, 0.011, 0.015, 0.025, 0.05, 0.1, 0.25, 0.5, 0.75, 0.9\}
	\item $\kappa$: \{1/391, 4/391, 10/391, 40/391, 0.25, 1, 2.5, 10, 25, 50, 100, 200\}
	\item \textit{n}: \{1, 3, 5\}
\end{itemize} 

\begin{figure}[t]
	\centering
	\includegraphics[width=0.9\columnwidth]{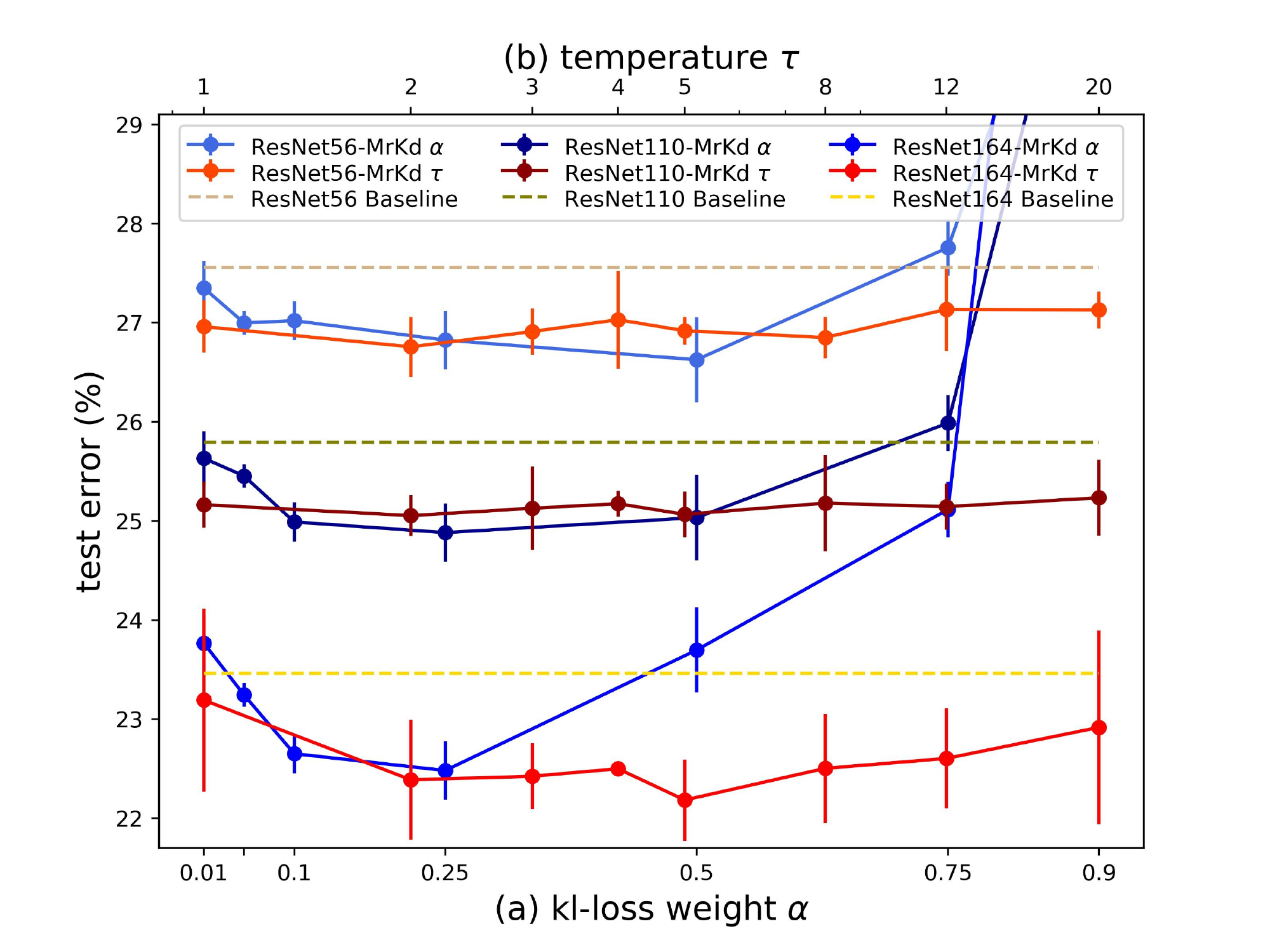} 
	\caption{Graph of Test Error v/s (a) KD loss weight $\alpha$ and (b) temperature $\tau$ of ResNet models on CIFAR100 dataset}
	\label{fig2}
\end{figure}

\begin{figure}[t]
	\centering
	\includegraphics[width=0.9\columnwidth]{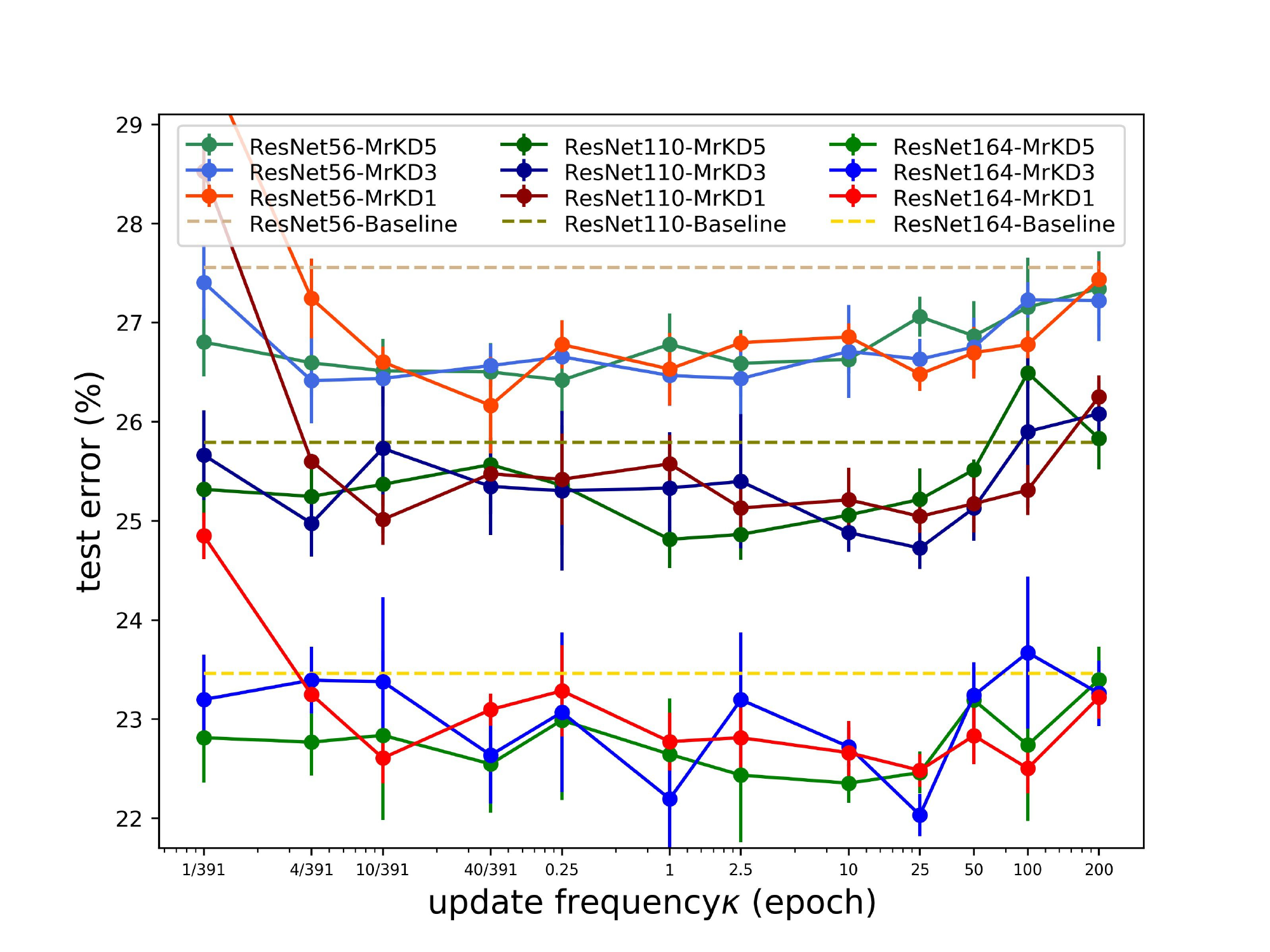} 
	\caption{Graph of Test Error v/s update interval $\kappa$ of ResNet models on CIFAR100 dataset}
	\label{fig3}
\end{figure}
Note that the unit of $\kappa$ is epoch. As the batch size we set is 128, the total iterations of an epoch are 391; thus, the $\kappa$=10/391 means we update the copies every 10 steps, and  $\kappa$=200 means that we never update the copies during the 200 epochs training. The control variates method is used below to show the result, which means that we set other hyper-parameters to the optimal value except for the one we want to evaluate.

\subsubsection{KL-Loss Weight $\alpha$ and Temperature $\tau$}
As $\alpha$ and $\tau$ are relatively independent, we evaluate them on MrKD with copy amount 1 by a simple line search. The best values are \{$\alpha$:0.25, $\tau$:3\}. 
As the blue lines show in Fig. 2(a), the $\alpha$ is relatively smaller and more sensitive than traditional distillation methods. We argue that since the model backup can be much worse than the current model, the KD loss should not guide the model learning as a strong leader but should act as a reference with a lower $\alpha$.

\cite{Hinton2015} shows that if the teacher and student are similar in size, the temperatures above 8 can give similar results. Furthermore, when the student model is much smaller, the best temperature reduces to 2.5 to 4  to make the KD loss harder and focus on matching larger logit values. As the copies and the current model have identical structures, the flat red lines in Figure 2(b) indicate that MrKD achieves a similar result with $\tau$ in the range [2, 8]. Compared to shallower models, ResNet164 is more sensitive to these two parameters.

For LsrKD, after grid search on ResNet-164 and WRN-40-4, we got error rate lines similar to Fig.2 with optimal value \{$\alpha$:0.1, $\tau$:3\}. The temperature $\tau$ is identical to MrKD. The KL loss ratio $\alpha$ is the same as the traditional LSR method and smaller than MrKD because the uniform distribution teacher offers less information than a real teacher.  From the dash lines in Fig. 4, we can see that LsrKD with $\tau$=1, which is equivalent to standard LSR method, gets worse results than baselines on ResNet-164 and WRN-40-4, whereas the improvement occurs when applying a softer temperature $\tau$=3.

\subsubsection{Model Backup Frequency $\kappa$ and Copy Amount \textit{n}}

The update frequency $\kappa$ and copy amount \textit{n} for MrKD are correlated since the backups' total epoch span is $\kappa$*\textit{n}.

In Fig. 3, we can see that if the step interval of the model backups is quite small, the accuracy will drop because the copy is too similar to the current model, then the regularization will not be helpful and may stumble the current model from learning. On the other hand, if the step is too large, the copies will be worse and lagging, then MrKD will also mislead and destabilize the learning. In conclusion, two ambivalent factors that influence the performance of MrKD while $\kappa$ changing: accuracy and diversity. For high accuracy, we need to be updated the copies frequently, while for diversity, the copies need to be far from the current model. 

The shallower model (ResNet56) is relatively insensitive to $\kappa$. On ResNet110 and ResNet164, we can see clearly in Fig. 3 that there are multiple local optimal  $\kappa$ values due to the two factors' non-linear mutual effect. The optimum values are  (\textit{n}, $\kappa$) $\in$ \{ (1, 25), (3, 25), (5, 2.5)\}. The short standard deviation bars indicate that the optimal values are very stable. These optimal values of $\kappa$ are out of our expectation because updating copies every 25 epochs means more than 10\% raise of training error than the current model. The large update step interval indicates that diversity is more important than accuracy.

We named our training method MrKD-1, MrKD-3, MrKD-5 for different \textit{n} (=1,3,5). Fig. 3 shows a similar trend as DML: with more teachers participate in the knowledge distillation, the students perform better and more stable with less local optimum values - the three green lines for MrKD-5 are flatter and lower. On the other hand, since there are fewer peak values, the optimal error rate for MrKD-5 is slightly worse than MrKD3 on ResNet models.

\subsubsection{Teacher Correction Factor $\gamma$}
In Fig. 4, the baselines for TC methods are LsrKD: $\tau$=3 and MrKD-3, respectively. For LsrKD, the TC method gets some improvement on both networks when $\gamma$ is in the range [0.11, 0.25].
With the comparison of LsrKD-TC and MrKD-3-TC, we can see that a teacher with dark knowledge can significantly help training than a lifeless handcrafted teacher when they have the same maximum probability in the correct class. 

On the other hand, for MrKD, the TC method is not always helpful: in ResNet164, MrKD-3-TC is worse than MrKD-3, whereas in WRN-40-4, TC improves MrKD-3 only at a few values of $\gamma$. This observation shows that a redesigned teacher with a constant value on the correct class is not always better than the original nature distribution with less accuracy. 

\begin{table*}[t]
	\caption{Result of our methods on CIFAR100 dataset}\smallskip
	\centering
	\smallskip\begin{tabular}{l|c|c|l|l|l|l|l|l|l}
		\textbf{Model}&\textbf{ \#params} & \textbf{Baseline} & \textbf{LSR} & \textbf{LsrKD}& \textbf{LsrKD-TC}& \textbf{MrKD1} & \textbf{MrKD3}& \textbf{MrKD3-TC}\\ 
		ResNet20 & 0.3M & 30.93  & \underline{31.06} & \underline{30.98}& \underline{31.00} (±0.14)& 30.83$\downarrow$  & \textbf{30.56}$\downarrow$ &  \underline{30.99} (±0.40)\\
		ResNet32& 0.5M & 29.49  & 28.93$\downarrow$   & 29.10& 29.16 (±0.16)&  \textbf{28.56}$\downarrow$ & 28.85 &28.74 (±0.41)\\
		ResNet44& 0.7M & 28.38  & 27.58$\downarrow$  & 27.92& 27.80 (±0.22) $\downarrow$ &  27.59$\downarrow$  & \textbf{27.37}$\downarrow$&27.43 (±0.10)\\
		ResNet56& 0.9M & 27.56 & 27.12$\downarrow$   & 27.34&  27.33 (±0.25)& \textbf{26.48}$\downarrow$ & 26.59&26.96 (±0.25)\\
		ResNet110& 1.7M & 25.79  & 25.48$\downarrow$  & 25.60 &25.53 (±0.43) & 25.05$\downarrow$ & \textbf{24.73}$\downarrow$ &25.25 (±0.16)\\
		PreResNet110& 1.7M & 25.63  & 25.49$\downarrow$   & 25.42$\downarrow$& 25.52 (±0.09)& 25.03$\downarrow$ & 24.93$\downarrow$ &\textbf{24.68} (±0.23)$\downarrow$ \\
		ResNet164& 1.7M & 23.46 & \underline{23.44} & 22.89$\downarrow$&22.51 (±0.30)$\downarrow$ & 22.62 & \textbf{22.23}$\downarrow$&22.56 (±0.50)\\
		PreResNet164& 1.7M & 22.04  & \underline{22.33} & 22.08$\downarrow$ & 22.04 (±0.28)$\downarrow$ & 21.87$\downarrow$ & \textbf{21.58}$\downarrow$& 21.59 (±0.32)\\
		WRN-40-4  & 9.0M& 20.73 & \underline{20.96}  & \underline{20.71}$\downarrow$ &20.52 (±0.20)$\downarrow$ & 20.21$\downarrow$ & 20.04$\downarrow$  & \textbf{19.82} (±0.23)$\downarrow$\\
		WRN-16-8 & 11.0M& 20.35 & 19.66$\downarrow$  & 19.88 & 19.98  (±0.18)& 20.25 & \textbf{19.63}$\downarrow$&19.84 (±0.13)\\
		WRN-28-10 & 36.5M & 19.00  & \underline{20.08} &  \underline{19.32}$\downarrow$ &\underline{19.20} (±0.33)$\downarrow$ &  18.94$\downarrow$ & \underline{19.01}& \textbf{18.51} (±0.33)$\downarrow$\\
		ResNeXt-29, 8×64d & 34.5M& 18.27  & \textbf{17.44}$\downarrow$ & 17.55&17.54 (±0.05) &  \underline{18.24} & 17.55$\downarrow$&17.63 (±0.21)\\
	\end{tabular}
	\label{table2}
\end{table*}
\subsubsection{Hyper-Parameters of CIFAR10 and CINIC10 datasets}

We set the TC factor $\gamma$ for LsrKD as 0.25 instead of 0.025, since there are only 10 classes in CIFAR10 and CINIC10.
We also found that setting the $\alpha$ for MrKD methods as 0.1 instead of 0.25 improves the results.
For CIFAR10 and CINIC10, we only adjust the critical hyper-parameters to keep universality. We believe that there would be better results on both datasets through a thorough search than we report in this paper.

\begin{figure}[t]
	\centering
	\includegraphics[width=0.9\columnwidth]{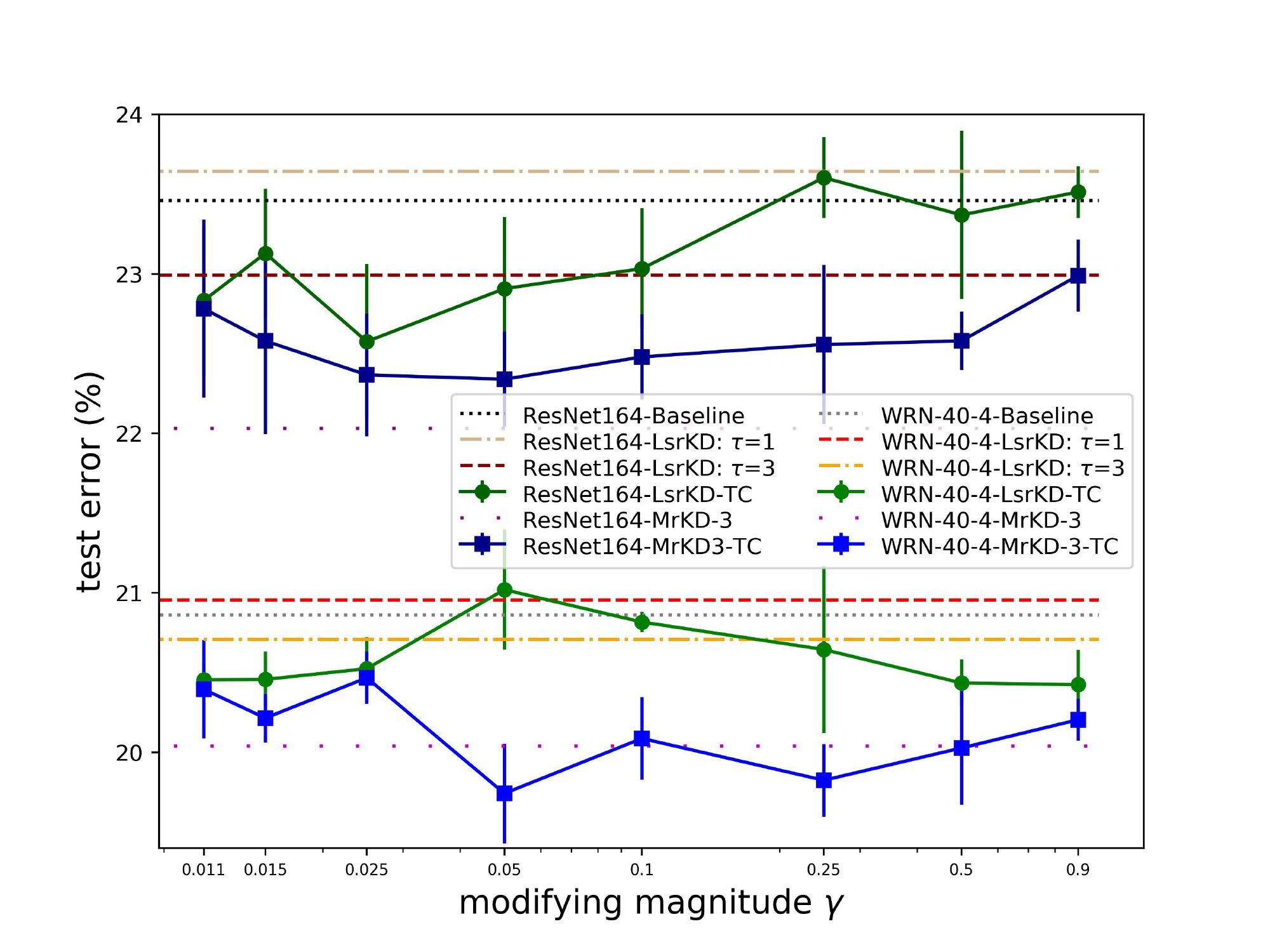} 
	\caption{Graph of Test Error v/s Teacher Correction factor $\gamma$ of ResNet models on CIFAR100 dataset}
	\label{fig4}
\end{figure}

	\begin{table*}[t]
	\caption{Result of our methods on CIFAR10 and CINIC10 dataset}\smallskip
	\centering
	\smallskip\begin{tabular}{l|l|l|l|l|l|l|l|l}
		\textbf{CIFAR10}\\
		\textbf{Model} & \textbf{Baseline} & \textbf{LSR} & \textbf{LsrKD}& \textbf{LsrKD-TC}& \textbf{MrKD1} & \textbf{MrKD3}& \textbf{MrKD3-TC}\\ 
		
		ResNet20 & 7.50  & 7.32$\downarrow$ & 7.21$\downarrow$ & 7.33 (±0.22) & \textbf{7.17}$\downarrow$  & 7.37 &  7.17 (±0.20)$\downarrow$\\
		ResNet32 & \textbf{6.21}  & \underline{6.48}   &  \underline{6.55} & \underline{6.37} (±0.12) $\downarrow$&  \underline{6.34} &  \underline{6.46} &\underline{6.23} (±0.06)$\downarrow$\\
		ResNet56 & 5.83 & \underline{5.98}   & \textbf{5.77}$\downarrow$&  \underline{5.84} (±0.08)& \underline{5.84} &  5.78$\downarrow$ &\underline{5.81} (±0.14)\\
		ResNet164 & 5.16 & \underline{5.22} & \underline{5.21}& 5.06 (±0.14)$\downarrow$ & \underline{5.21} & \textbf{4.83}$\downarrow$ &\underline{5.16} (±0.19)\\
		WRN-16-8 & 4.17 & 3.96$\downarrow$  & 3.88$\downarrow$ & 3.93 (±0.06)& 3.83 $\downarrow$& 3.81$\downarrow$ & \textbf{3.77} (±0.06)$\downarrow$\\
		
		\textbf{CINIC10}\\
		ResNet20 & 17.26  & 17.16$\downarrow$ &  \underline{17.36} & 17.14 (±0.14)$\downarrow$& 17.05$\downarrow$  & \textbf{16.88}$\downarrow$ &  17.07 (±0.06)\\
		ResNet32 & 16.13  & 15.92$\downarrow$   & 15.88$\downarrow$&  15.95 (±0.21)&  15.78$\downarrow$ & \textbf{15.71}$\downarrow$ &15.96 (±0.15)\\
		ResNet56 & 15.27 & \underline{15.39}   & 15.17$\downarrow$&  15.20 (±0.04)& 15.08$\downarrow$ & \textbf{14.82}$\downarrow$&15.23 (±0.10)\\
		ResNet164 & 13.41 & \underline{13.61} &  \underline{13.76}& \underline{13.41} (±0.23)$\downarrow$ & 13.21$\downarrow$ &\underline{13.53}& \textbf{13.10} (±0.22)$\downarrow$	\\
		WRN-16-8 & 11.52 & 11.22$\downarrow$  & 11.13$\downarrow$ & 11.13 (±0.03)& \textbf{11.04}$\downarrow$ & 10.90$\downarrow$& 11.05 (±0.05)\\
		
	\end{tabular}
	\label{table3}
\end{table*}

	\subsection{Results}
	Experimental results are shown in Table 1 and 2. The error rates without improvement comparing to the baselines are underlined, and the bold results are the best ones for every network. The down arrow indicates a lower error rate is obtained by the method than its predecessor at the previous column.
	\subsubsection{Results on CIFAR100}
	In Table 1, we can see that LSR is working well on shallow networks, while struggles to get improvement on deeper networks where LsrKD performs much better. 
	
	It can be observed that almost all the MrKD-1 results are better than LsrKD methods. MrKD-3 improves most of the results further, with an increment from 0.37\% to 1.2\% in CIFAR100 dataset. However, there is no improvement for WRN-28-10 on MrKD-3 and all previous methods. We suppose that for a few deep or complicated models that are more sensitive, the hand-crafted or inaccurate teacher of our methods may perturb the learning and offset the regulation effect. From the last column, we can see that MrKD-3-TC is the only method to improve this network.
	
	LsrKD with TC further improves on most deep networks, whereas fewer models benefit from TC for MrKD-3. We can infer that the MrKD-3 is good enough in most cases, while TC is essential to LsrKD.
	
	\subsubsection{Results on CIFAR10 and CINIC10}
	Table 2 shows similar improvements in Table 1. The LSR method only helps half on the networks with tiny improvements, whereas our methods perform better. Most networks benefit from a part of our methods.  
	
	Compared to Table 1, the more underlines in Table 2 indicate a worse performance on CIFAR10 and CINIC10. There are two reasons: (1) the networks distillate less information on 10-class datasets problems. (2) The gap between the test and training error rate on CIFAR10 and CINIC 10 is lower than on CIFAR100; then, KD methods’ generalization effect is not significant. 
	
	\section{Conclusion and Future work}
	
	In this paper, we propose simple but effective Knowledge Distillation methods without external knowledge or data. Specifically, with a softer temperature, LsrKD and its Teacher Correction variant are good substitutions of Label Smoothing Regularization. Adopting model parameter backup as the teacher of self-distillation, MrKD obtains better results than LsrKD. 
	
	To our best knowledge, in the supervised learning area, MrKD is the first method to try to utilize the model backup with a large step interval to the current model parameters. This mechanism is fascinating and can be combined with many existing KD method conveniently. Weight averaging can merge the current and the historical model to obtain a better teacher like Mean Teacher  \cite {Tarvainen2017}. Other knowledge may also be mined from the model backups by weights regularization, intermediate layer outputs, or attention maps, instead of the logits output. The effectiveness of Teacher Correction on other KD methods needs to be evaluated in the future.

	\bibliographystyle{aaai}
	\bibliography{ic, kd}

\end{document}